\documentclass[conference]{IEEEtran}
\IEEEoverridecommandlockouts
\usepackage{cite}
\usepackage{amsmath,amssymb,amsfonts}
\usepackage{algorithmic}
\usepackage{graphicx}
\usepackage{textcomp}
\usepackage{xcolor}

\def\BibTeX{{\rm B\kern-.05em{\sc i\kern-.025em b}\kern-.08em
    T\kern-.1667em\lower.7ex\hbox{E}\kern-.125emX}}

\usepackage{caption}
\usepackage{subcaption}

\usepackage[hidelinks]{hyperref}

\begin{document}

\title{Shared Autonomy through LLMs and Reinforcement Learning for Applications to Ship Hull Inspections
}
\author{
\IEEEauthorblockN{Cristiano Caissutti}
\IEEEauthorblockA{
\textit{Dept. of Information Engineering} \\
\textit{University of Pisa, Italy} \\
Pisa, Italy \\
cristiano.caissutti@phd.unipi.it}
\\
\IEEEauthorblockN{Paolo Marinelli}
\IEEEauthorblockA{
\textit{Dept. of Information Engineering} \\
\textit{University of Pisa, Italy} \\
Pisa, Italy \\
paolo.marinelli@ing.unipi.it}
\and
\IEEEauthorblockN{Estelle Gerbier}
\IEEEauthorblockA{
\textit{Dept. of Information Engineering} \\
\textit{University of Pisa, Italy} \\
Pisa, Italy \\
estelle.gerbier@ing.unipi.it}
\\
\IEEEauthorblockN{Andrea Munafo\textquotesingle}
\IEEEauthorblockA{
\textit{Dept. of Information Engineering} \\
\textit{University of Pisa, Italy} \\
Pisa, Italy \\
andrea.munafo@unipi.it}
\and
\IEEEauthorblockN{Ehsan Khorrambakht}
\IEEEauthorblockA{
\textit{Dept. of Information Engineering} \\
\textit{University of Pisa, Italy} \\
Pisa, Italy \\
ehsan.khorrambakht@ing.unipi.it}
\\
\IEEEauthorblockN{Andrea Caiti}
\IEEEauthorblockA{
\textit{Dept. of Information Engineering} \\
\textit{University of Pisa, Italy} \\
Pisa, Italy \\
andrea.caiti@unipi.it}
}

\maketitle


\begin{abstract}

Shared autonomy is a promising paradigm in robotic systems, particularly within the maritime domain, where complex, high-risk, and uncertain environments necessitate effective human-robot collaboration. This paper investigates the interaction of three complementary approaches to advance shared autonomy in heterogeneous marine robotic fleets: (i) the integration of Large Language Models (LLMs) to facilitate intuitive high-level task specification and support hull inspection missions, (ii) the implementation of human-in-the-loop interaction frameworks in multi-agent settings to enable adaptive and intent-aware coordination, and (iii) the development of a modular Mission Manager based on Behavior Trees to provide interpretable and flexible mission control. Preliminary results from simulation and real-world lake-like environments demonstrate the potential of this multi-layered architecture to reduce operator cognitive load, enhance transparency, and improve adaptive behaviour alignment with human intent. Ongoing work focuses on fully integrating these components, refining coordination mechanisms, and validating the system in operational port scenarios. This study contributes to establishing a modular and scalable foundation for trustworthy, human-collaborative autonomy in safety-critical maritime robotics applications.
\end{abstract}




\section{Introduction}\label{sec:introduction}
In recent years, the increasing deployment of autonomous marine vehicles in complex environments such as ports, coastal infrastructures, and critical maritime assets has highlighted the necessity of creating trustworthy decision-making frameworks where human-robot collaboration is improved. These environments are dynamic, cluttered, and only partially structured, making full autonomy challenging and potentially unsafe due to sensor failures, communication loss, or unpredictable behaviours.
In this context, trustworthiness goes beyond system robustness; it requires autonomous systems to be intelligible, predictable, and responsive to human oversight. Shared autonomy emerges as a promising approach, enabling robots to operate independently while integrating human intervention when needed.


In port environments, for instance, autonomous vehicles must navigate safely around human personnel, vessels, and critical infrastructure, all while accomplishing mission objectives such as hull inspection, pollution monitoring, and other tasks. Yet, autonomous systems still struggle with edge cases and rare events that require human judgment. Shared autonomy, therefore, emerges as a promising solution that enables autonomous systems to operate, but sharing control or seeking assistance when uncertainty grows or human input is more suitable.




This work aims to advance the development of trustworthy shared autonomy frameworks for heterogeneous marine robotic fleets operating in complex, safety-critical environments. 
Trustworthiness here is understood not only as system robustness, but as the ability of the autonomous system to remain intelligible, predictable, and adaptable in the presence of human supervision and intervention. 

To address these challenges, this work proposes a hybrid architecture for shared autonomy in heterogeneous marine fleets. The system integrates natural language interfaces powered by large language models (LLMs) \cite{brown2020language}, modular mission planning through Behavior Trees (BTs)\cite{8729810}, and learning-based execution by Deep Reinforcement Learning (DRL)\cite{reddy2018sharedautonomydeepreinforcement}. These components allow the operator to specify high-level goals, monitor mission progress, and rely on agents that adapt their behaviour based on both context and human intent.


This work provides a preliminary validation of this approach through a hybrid evaluation strategy. The LLM integration has been tested in a simulated environment, allowing an operator to specify high-level goals and oversee mission status through natural language, while the RL component has been deployed on a real marine robot in a lake-like setting, demonstrating the feasibility of adaptive behaviour in real-world conditions. The full integration of the BT mission manager is still under development; these initial results support the core design assumptions of the architecture.

The main contribution of this work lies in the formulation of a shared autonomy framework that integrates interpretability, adaptability, and human oversight through a modular and layered architecture. By combining language-based interaction, structured mission logic, and learning-driven execution, this approach demonstrates a practical and scalable framework for human-guided autonomy in multi-agent maritime tasks, validated in a representative hull inspection scenario.

The remainder of this paper is organized as follows. Section 2 reviews related work across language-based interfaces, behaviour planning, and learning-based autonomy. Section 3 presents the architecture in detail, with a focus on human interaction and mission control. Section 4 describes a hull inspection scenario used for evaluation. Finally, Section 5 concludes the paper and outlines future research directions.


\section{Background}\label{sec:detect_and_class}
Over the past decade, the field of underwater robotics has begun to converge with broader developments in autonomous systems. While significant progress in shared autonomy, reinforcement learning, and human-robot interaction has traditionally originated from the aerial, terrestrial, and service robotics domains~\cite{Kemp2007, Abbeel2010, Yoon2018}, marine systems have been rapidly catching up~\cite {Caiti2019, Fallon2021}. This shift has been enabled by advances in simulation environments, which allow for extensive and cost-effective training of underwater robots in realistic scenarios \cite{cieslak2019stonefish, gentili2024amulti}, improved communication infrastructures, which help mitigate the traditional limitations of underwater data transmission and enable more effective human-robot collaboration and the adoption of modular autonomy frameworks, which facilitate flexible integration and rapid adaptation of autonomous capabilities in complex underwater settings.
As a result, tools and techniques that were once confined to ground and aerial domains, such as behaviour trees, learning-based control, and natural language interfaces, are increasingly being explored in underwater applications. This growing convergence is gradually dissolving the boundaries between underwater and mainstream robotics, encouraging the transfer and adaptation of established methods to address the unique challenges of the maritime environment. In this context, new perspectives are emerging on trust, adaptability, and resilience, particularly in domains characterised by harsh conditions, limited communication, and high operational risk. This work proposes an architecture that integrates elements from modern robotics, such as interpretable planning, coordinated learning, and intuitive human supervision, tailored to the specific needs of multi-agent marine systems. Among recent developments with promising potential is the use of LLMs to mediate interaction between humans and autonomous agents \cite{khorrambakht2025}. LLMs can translate natural language instructions into structured commands, easing communication and reducing the cognitive burden on operators \cite{kannan2024smart}. This allows for new possibilities for human supervision, especially in scenarios where expert input is required at high levels of abstraction. Projects such as SayCan and Code-as-Policies (\cite{ahn2022can, liang2023code}) have demonstrated the ability of these models to connect language with action planning. However, their integration into real-world systems remains limited by issues of hallucination, lack of grounding in real-time sensor data, and general unpredictability. In the proposed approach, LLMs are therefore used not as autonomous decision makers but as high-level interfaces for goal specification and feedback summarisation, always in coordination with more grounded control mechanisms. To manage the logic of mission execution in a structured yet flexible way, we adopt BTs as our core decision-making framework.
Behavior Trees, originally developed in the domain of game artificial intelligence \cite{ogren2012increasing}, have been increasingly adopted in robotics due to their modularity, transparency, and runtime flexibility. Their hierarchical structure allows complex missions to be constructed from simple, reusable components that can be monitored, tested, and modified by human operators. BTs have emerged as a state-of-the-art approach for mission execution in autonomous systems, offering operational robustness and adaptability that have been extensively validated across diverse domains, including robotics. In maritime robotics, BTs represent a notable advancement, enabling autonomous agents to perform complex, context-aware behaviours in highly dynamic and unpredictable environments. BTs offer a transparent decision-making process, making them particularly well-suited for human oversight, safety-critical operations, and coordinated multi-vehicle missions.
At the lowest level of autonomy, RL provides a means for agents to adapt their behaviour through interaction with the environment. In shared autonomy scenarios, RL is often combined with human guidance—through demonstrations \cite{reddy2018sharedautonomydeepreinforcement, christiano2023deepreinforcementlearninghuman}, corrections, or indirect feedback—enabling agents to learn not only how to perform a task, but also how to align their actions with human intent. While most prior work has focused on single-agent settings or assistive systems, the application of shared autonomy to multi-agent marine environments remains relatively underexplored. In our previous work \cite{khorrambakht2025}, we proposed a shared autonomy framework in which one vehicle is teleoperated, while the remaining agents are trained via DRL to observe and adapt to its behaviour. This configuration allows for emergent cooperation that is both scalable and intent-aware: the human-guided agent implicitly communicates mission priorities, and the rest of the fleet dynamically coordinates in response. The present architecture builds upon this foundation, generalising the model to support structured mission planning and transparent human oversight across all levels of autonomy. In our work, we address this gap by introducing a coordination model in which RL-trained agents adjust their behaviour based on the actions of a human-guided leader, supporting emergent, intent-aware cooperation within the fleet. Altogether, these components structure the basis of our proposal for a trustworthy shared autonomy architecture. In underwater robotics, trustworthiness extends far beyond robustness: it implies transparency, predictability, and the capacity for human intervention when needed. Traditional fault-tolerant control approaches have addressed robustness in isolation, but few systems provide the operator with a coherent, layered interface that integrates intent specification, interpretable logic, and adaptive behaviour. 

\begin{figure}
    \centering
    \includegraphics[width=1.0\linewidth]{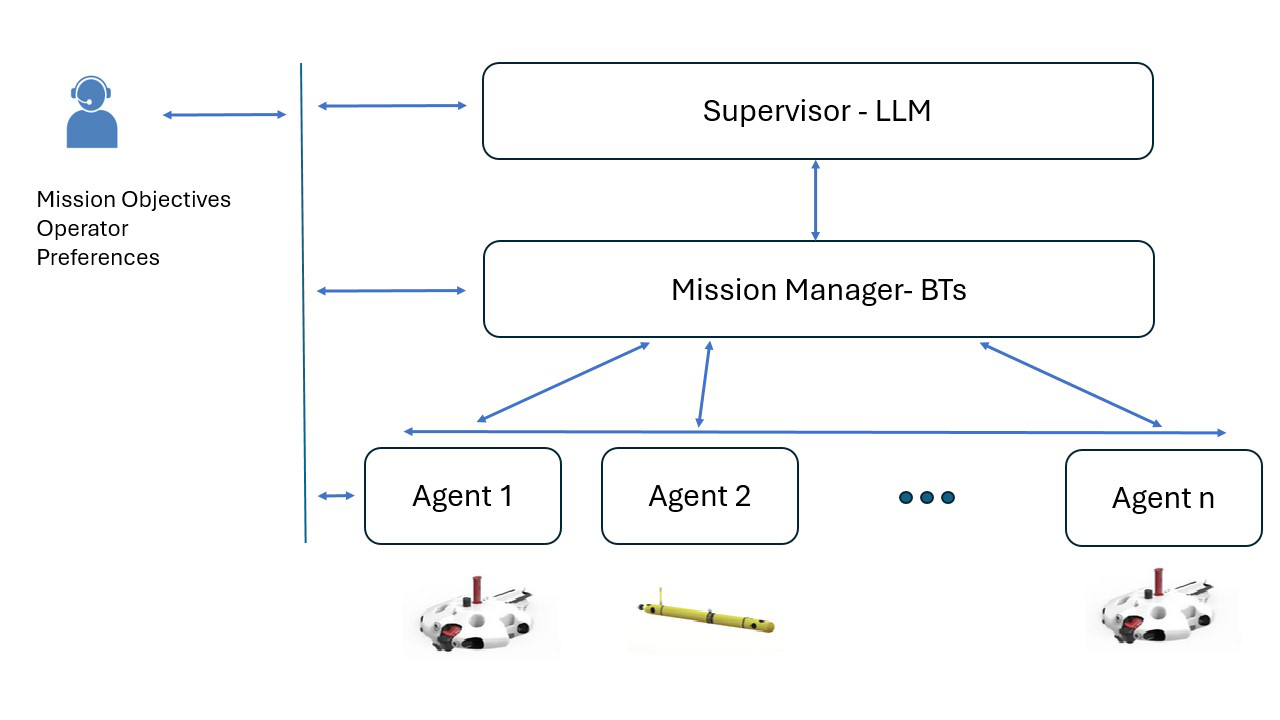}
    \caption{LLM-based Multi-Agent Framework: The operator interacts with the supervisory agent, who interprets the operator’s objectives into executable commands for the available robots. Each robot is part of the system through an agentic representative that reflects its status and capabilities.}
    \label{fig:llm-label}
\end{figure}

\section{Methods}\label{sec:results}

The proposed architecture is structured in three layers: (i) a natural language interface powered by an LLM, (ii) a modular mission manager based on BTs, and (iii) a multi-agent execution layer using DRL. 


Figure 1 illustrates the overall information flow across the system, where intent flows top-down from the operator to the agents, and feedback is propagated upward to support supervision and adaptation. Crucially, this layered design enables the human operator to access and influence every level of the system’s complexity. At the top layer, goals can be expressed in natural language and translated into structured tasks. At the mission level, operators can inspect, modify, or override the behavior tree that governs task execution. At the execution layer, they can monitor or operate on individual agents. This structure ensures that the operator remains in the loop not only as a supervisor but as an active partner, capable of interacting with both what the system is doing and how it is doing it. This renders the architecture inherently user-oriented, as it is conceived from the outset to uphold transparency, adaptability, and user agency throughout all components.
The architecture is instantiated through three core components, each responsible for a distinct level of system functionality: a language-based interface, a mission planning framework, and a multi-agent execution layer. 

\subsection{ Natural Language Interface with LLMs}
The first layer of the system enables goal specification and mission-level communication through a natural language interface powered by an LLM. In our implementation, we used the LLAMA-based model \cite{touvron2023llama} (Meta AI) as the underlying LLM, due to its open-access architecture and suitability for lighetweight inference tasks. The LLM is used to: \begin{itemize}
    \item Interpret human-issued instructions.
    \item Generate structured plans or parameters.
    \item Summarize feedback for the operator during and after mission execution.
\end{itemize}
The components were tested in simulation, using task templates and controlled queries to validate their capacity to parse high-level goals and convert them into actionable commands. 


\subsection{Behavior Trees for Mission Logic and Execution (in development)}
The second layer centers on a modular mission manager structured with Behavior Trees. This component is designed to execute structured plans (e.g., go to specified waypoints, inspection sequences, or emergency procedures) by an interpretable and reconfigurable execution graph. BTs offer multiple advantages in this context, including runtime reactivity, ease of debugging, and transparency of decision flow.
Although the full integration of BTs is still under testing, this work has defined the initial BT structures and developed mission templates aligned with key use cases (e.g., hull inspection, coordinated survey). 

\subsection{Multi-Agent Execution via Shared Autonomy and DRL}
The execution layer adopts a shared autonomy scheme in a leader-follower configuration. A single AUV—the leader—is teleoperated by a human operator, while the second AUV—the follower—is trained via deep reinforcement learning to adapt its behaviour based on the leader’s position. The leader communicates its pose at fixed intervals, and the follower adjusts its trajectory to maintain a predefined formation, while also considering proximity to the hull and obstacle avoidance.

This setup was tested in a hull inspection scenario, where the fleet operates near the ship hull to detect potential anomalies. Although only two agents are deployed in this study, the approach is designed to scale to larger fleets by extending the formation model and reusing the learned coordination policy. This structure enables a form of indirect shared autonomy, in which operator intent influences the fleet through the leader’s behaviour, ensuring coherent and adaptive group responses with minimal intervention.\\

Evaluation was performed using a hybrid setup combining a real-world deployment of the Zeno AUV \ref{fig:zeno} in a lake and a simulated second agent with equivalent dynamics. Simulated acoustic communication was established between the two agents, and a virtual ship hull was introduced to replicate a constrained inspection environment.

\section{Case Study: Intent-Aware Coordination in a Port Scenario}
\begin{figure}
    \centering
    \includegraphics[width=0.7\linewidth]{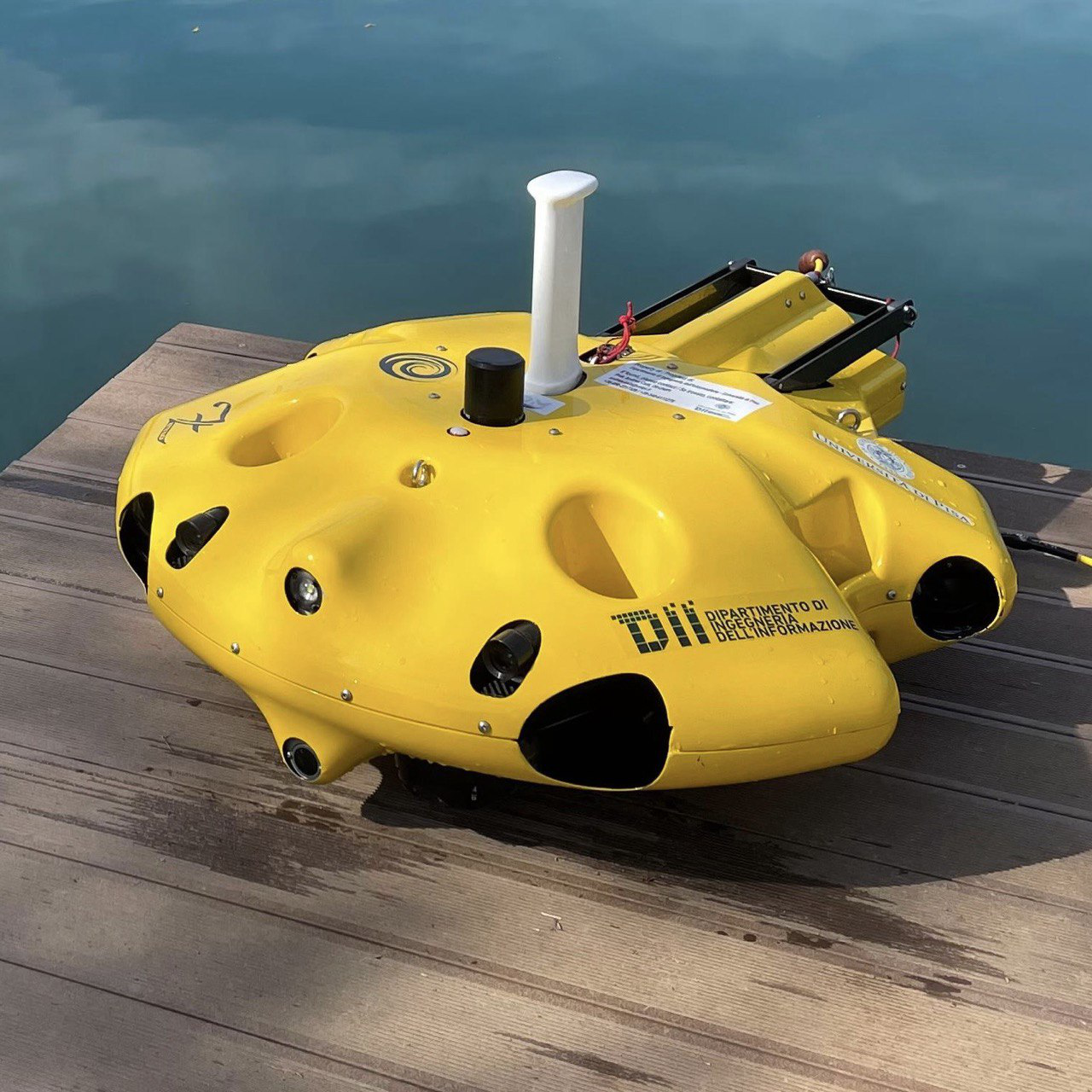}
    \caption{Zeno AUV, owned by CrossLab at the Department of Information Engineering, University of Pisa, Italy.}
    \label{fig:zeno}
\end{figure}

The case study involved a hybrid setup: one Zeno MDM AUV, shown in Fig. \ref{fig:zeno}, was deployed in a lake real-world environment, while a second agent, simulated in ROS using the same dynamic model, operated in parallel. The communication was simulated within the ROS architecture, without explicitly modeling the physical acoustic channel. A virtual ship hull was also included in the simulated environment to replicate the inspection scenario in constrained maritime settings.

The simulated experiments were conducted in a ROS-based environment using the dynamics of the Zeno MDM vehicle model. The environment includes physics-based dynamics but does not explicitly model underwater currents or communication noise. Similarly, sensor inputs such as camera feeds or distance measurements are simulated with ideal conditions, without introducing perception uncertainty or sensor noise. In particular, the follower sensing is modeled as a multibeam sonar with a wide field of view, used to track the PoI and avoid collisions with the leader or the hull.

\begin{figure}
    \centering
    \includegraphics[width=0.8\linewidth]{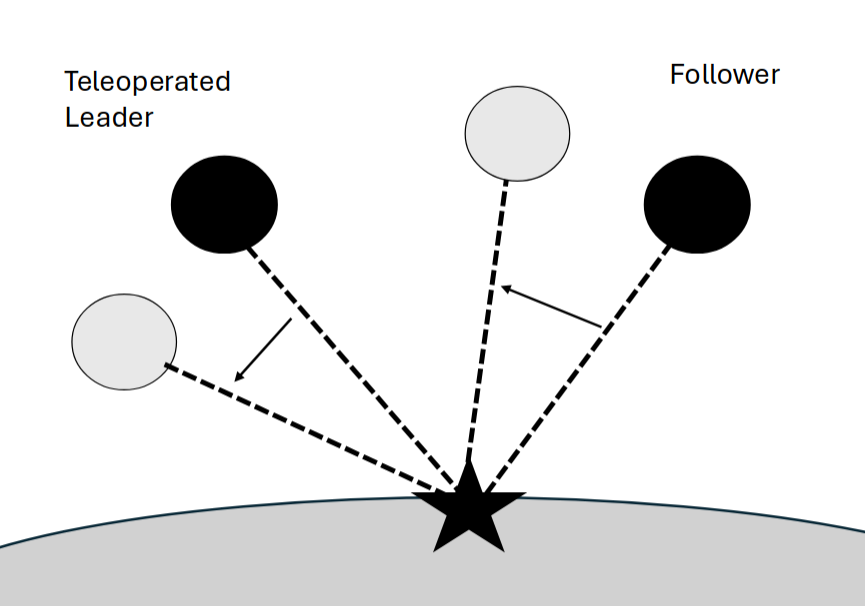}
    \caption{Schematic of the leader–follower shared autonomy. }
    \label{fig:lead-follow-label}
\end{figure}

\begin{figure}
    \centering
    \includegraphics[width=0.8\linewidth]{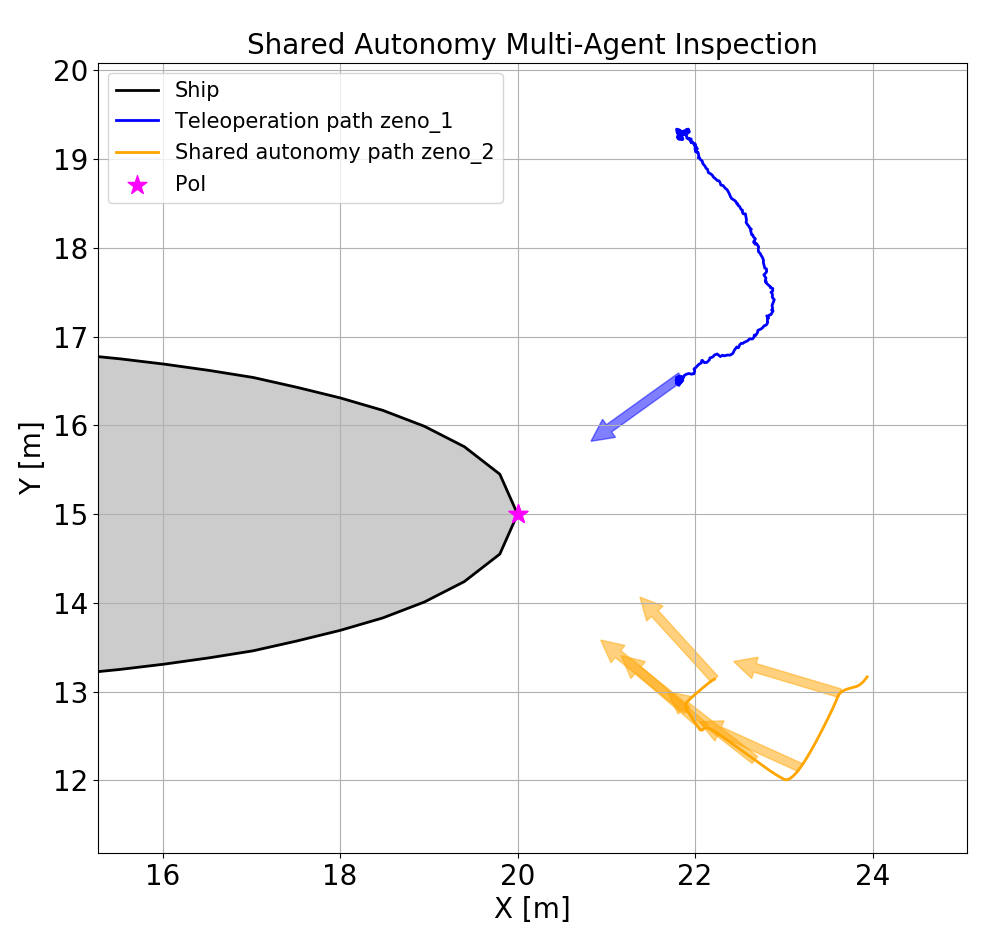}
    \caption{The trajectory of the fleet during the teleoperation.}
    \label{fig:enter-label-shared}
\end{figure}

A fleet of two agents was deployed:
\begin{itemize}
    \item A leader, controlled by a human operator through a joystick;
    \item A follower, trained via DRL to maintain a fixed spatial formation and ensure visibility of the Point of Interest (PoI) in the inspected hull, defined by the operator.
\end{itemize}
When the human operator takes control via the joystick, the leader switches from autonomous to manual mode. The follower dynamically adapts its motion based on the updated leader pose, preserving formation and maintaining continuous visibility of the PoI within its sensor footprint, which in our setup is modeled as a simulated forward-facing multibeam sonar. This behaviour is not scripted: the follower uses a policy learned via DRL that reasons over both relative positioning and sensory input. The resulting actions ensure that the PoI remains observable while avoiding collisions with static obstacles and with the leader itself.
The operator communicates intent via keyboard using natural language commands, which are processed by an LLAMA-based model. These are translated into structured tasks that drive the mission logic. Although the BT module is under active development, for this study, the mission phases were manually instantiated to emulate a BT-based structure.
Fig.\ref{fig:lead-follow-label} illustrates this coordination pattern. The black circles represent the current positions of the vehicles, both oriented toward the PoI (black star). The grey circles indicate the target configuration after the leader moves, with the follower preserving its position relative to the PoI. 
In this context, the reward function used to train the follower encodes multiple objectives: 
\begin{itemize}
    \item minimizing the distance to the desired formation point
    \item maintaining the PoI within the sensor field of view
    \item penalizing proximity to obstacles or the hull, including the leader
\end{itemize}
While a formal data-driven evaluation is planned for future work, the following performance metrics have been defined to assess coordination quality and mission robustness:
\begin{itemize}
    \item percentage of time the PoI remains within the follower's field of view
    \item average deviation from the ideal formation point,
    \item number of safety violations or near-collision events.
\end{itemize}
These metrics are intended to quantify the robustness of the follower’s policy in maintaining formation, preserving target visibility, and ensuring safe coordination during hull inspection scenarios.
Preliminary qualitative results, illustrated in Fig.\ref{fig:enter-label-shared}, suggest that the learned policy enables the follower to maintain the required formation and keep the PoI within its sensor footprint, while dynamically avoiding both static obstacles and the leader's trajectory. This result is coherent fleet behaviour that reflects human intent while preserving safety and mission coverage.

\begin{figure*}[h!]
    \centering
    \includegraphics[width=1.0\linewidth]{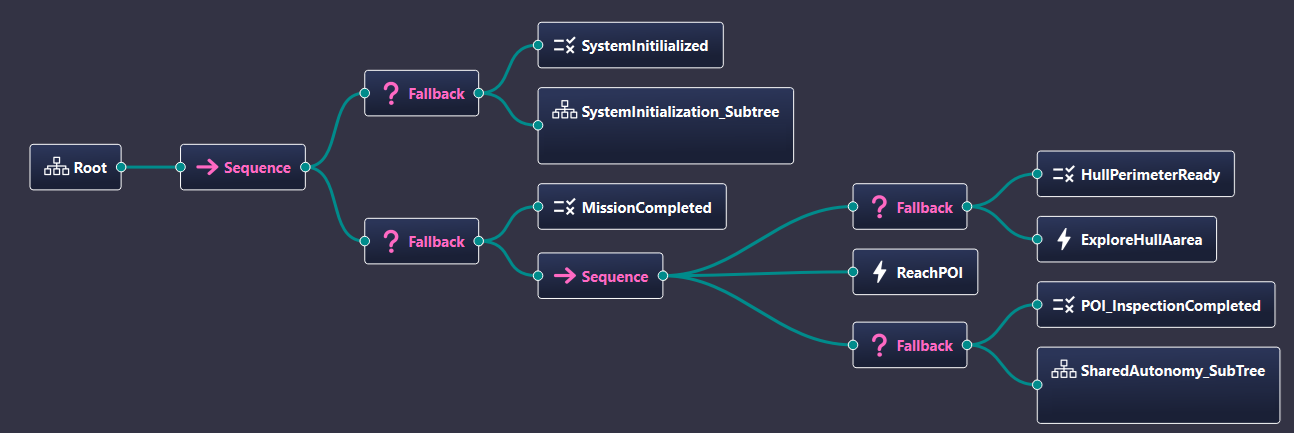}
    \caption{Behavior Tree mission structure.}
    \label{fig:enter-labelb}
\end{figure*}

The simulation environment included physics-based dynamics and synthetic sensing. At the high level, a natural language interface powered by an LLM enabled the operator to specify goals like “inspect the port side of the hull” or “report anomalies near the stern”, which were parsed into structured mission tasks. Although the behavior tree module is still under active development, for this study the mission logic was manually instantiated to emulate BT-based planning and allow supervision over task execution phases (see \ref{fig:enter-labelb} for a schematic representation).


These results highlight the feasibility and potential of combining LLM-based interaction, learning-based coordination, and modular mission management to support trustworthy, human-centered autonomy in port environments.

These experimental insights support the architectural vision of shared autonomy as a layered, modular framework capable of aligning robotic behaviour with human supervision in dynamic maritime operations.


\section{Conclusion and Future Works}\label{sec:conclusion}
This paper has presented a preliminary shared autonomy framework for heterogeneous marine robotic fleets operating in complex and safety-critical maritime environments. The architecture integrates a natural language interface based on LLMs, a modular mission management system utilizing BTs, and a multi-agent execution layer employing DRLs. This integration aims to enable human-robot collaboration characterized by transparency, predictability, and adaptability.

Initial validation in both simulated and real-world lake-like environments indicates that the natural language interface effectively supports high-level goal specification and reduces operator cognitive load. The reinforcement learning agents demonstrated dynamic adaptation to human-guided behaviours, while the modular mission management framework shows potential for interpretable and flexible mission execution.

Although the integration of the BT component remains ongoing, preliminary results demonstrate the feasibility of the overall architectural approach. The framework establishes a foundation for trustworthy, human-collaborative autonomy in multi-agent marine robotic systems; however, further development and rigorous evaluation are necessary.

Future work will prioritize the comprehensive integration of the mission planning layer, enhancement of coordination mechanisms among agents, and extensive validation in operational port environments. Additionally, efforts will focus on grounding language models more effectively in real-time sensor data to improve system reliability and operator trust.

Collectively, these findings constitute an initial step toward advancing shared autonomy methodologies in marine robotics by providing a modular and scalable framework applicable to safety-critical maritime operations.
\section*{Acknowledgments}
The research leading to these results has received funding from
Project ”COMET: Composable autonomy in Marine Environments” CUP:
2022TPSX25 and the FORELab (Future ORiented Engineering Laboratory) project, funded by the University of Pisa under the framework of strategic research initiatives.

\bibliographystyle{IEEEtran}
\bibliography{oceans}

\end{document}